\documentclass[12pt, peerreviewca, onecolumn]{IEEEtran}
\usepackage{graphicx,amsmath,amsfonts,amssymb,epsfig}
\usepackage{pseudocode}
\usepackage{subfigure}
\usepackage{algpseudocode}
\newtheorem{lemma}{\hskip\parindent\bf{Lemma}}[section]
\newtheorem{proposition}{\hskip\parindent\bf{Proposition}}[section]
\newtheorem{theorem}{Theorem}

\begin{document}
\title{GNCGCP - Graduated NonConvexity and Graduated Concavity Procedure
\thanks{}}
\author{Zhi-Yong Liu, Member IEEE, and Hong Qiao, Senior Member, IEEE
\thanks{Z. Y. Liu and Hong Qiao are with the State Key Laboratory of Management and Control for Complex Systems,
Institute of Automation, Chinese Academy of Sciences, Beijing, China. Corresponding author: Z.Y. Liu (zhiyong.liu@ia.ac.cn)}} \maketitle

\begin{abstract}
In this paper we propose the Graduated NonConvexity and Graduated Concavity Procedure (GNCGCP) as a general optimization
framework to approximately solve the combinatorial optimization problems on the set of partial permutation matrices. GNCGCP
comprises two sub-procedures, graduated nonconvexity (GNC) which realizes a convex relaxation and graduated concavity (GC) which
realizes a concave relaxation. It is proved that GNCGCP realizes exactly a type of convex-concave relaxation procedure (CCRP),
but with a much simpler formulation without needing convex or concave relaxation in an explicit way. Actually, GNCGCP involves
only the gradient of the objective function and is therefore very easy to use in practical applications. Two typical NP-hard
problems, (sub)graph matching and quadratic assignment problem (QAP), are employed to demonstrate its simplicity and
state-of-the-art performance.

\end{abstract}
\section{Introduction}
The recently proposed Path following and extended Path following algorithms exhibited state-of-the-art performances \cite{Zaslavskiy09,zyliu12} on
equal-sized graph matching problems. As a typical NP-hard problem, equal-sized (with size $N$) graph matching under the one-to-one constraint can be
formulated as follows,
\begin{equation}\label{eqn_gm}
\min_\mathbf{X} F(\mathbf{X}),\hspace{0.5cm}\mathrm{ s.t.} \mathbf{X}\in\mathcal{P}, \mathcal{P}:=\{\mathbf{X}|\mathbf{X}_{ij}=\{0,
1\},\sum_{j=1}^N{\mathbf{X}_{ij}}=1, \sum_{i=1}^N{\mathbf{X}_{ij}}=1,\forall i, j\},
\end{equation}
where $\mathcal{P}$ denotes the set of ($N\times N$) permutation matrices, $F(\mathbf{X})$ is given later by (\ref{subgm-obj}) or (\ref{gm_obj}).  By
relaxing $\mathcal{P}$ to its convex hull, i.e., the set of ($N\times N$) doubly stochastic matrices denoted by $\mathcal{D}$, the (extended) Path
following algorithm proposed the convex and concave relaxation procedure (CCRP) formulated by a weighted linear combination of a convex relaxation
$F_v(\mathbf{X})$ and a concave relaxation $F_c(\mathbf{X})$ of $F(\mathbf{X})$ as follows \cite{Zaslavskiy09,zyliu12,zyliu12-2},
\begin{equation}\label{ccrp}
F_{\eta}(\mathbf{X}) = (1-\eta)F_v(\mathbf{X})+\eta F_c(\mathbf{X}), \mathbf{X}\in\mathcal{D}.
\end{equation}
In implementation, $\eta$ increases gradually from 0 to 1, making $F_{\eta}(\mathbf{X})$ become gradually from $F_v(\mathbf{X})$ to
$F_c(\mathbf{X})$, whose minima locate exactly in $\mathcal{P}$. Similar to the graduated nonconvexity \cite{Blake87} algorithm, the (extended) Path
following is a deterministic annealing method which usually finds a good suboptimal solution, but at a much less computational cost than stochastic
simulated annealing techniques. As a state-of-the-art optimization algorithm, the (extended) Path following showed superior performance but has
difficulties in finding convex or concave relaxation, which thus greatly hinders its practical applications. For instance, neither the convex nor
concave relaxation proposed by (extended) Path following \cite{Zaslavskiy09,zyliu12} on equal-sized graph matching is applicable on the subgraph
matching defined on the set of partial permutation matrices.

In this paper we will propose the Graduated NonConvexity and Graduated Concavity Procedure (GNCGCP) to equivalently realize the
CCRP in (\ref{ccrp}) on \emph{partial} permutation matrix (with permutation matrix as a special case), but in a much simpler way
without involving the convex or concave relaxation explicitly. Actually, GNCGCP needs only the gradient of the objective
function, making it very easy to use in practical applications. Two case studies on (sub)graph matching and quadratic assignment
problem (QAP) witness the simplicity and state-of-the-art performance of GNCGCP. The GNCGCP is proposed in the next Section,
followed by the (sub)graph matching and QAP problems discussed in Sections \ref{s3} and \ref{s4} respectively, and finally
Section \ref{s5} concludes this paper.

\section{Graduated NonConvexity and Graduated Concavity Procedure}\label{s2}
\subsection{Formulation and Algorithm}
In this paper we consider the optimizations on the set of ($M\times N$) partial permutation matrices $\Pi$ as,
\begin{equation}\label{oriobj}
\min_X F(\mathbf{X}),\hspace{0.5cm} \mathrm{s.t.} \mathbf{X}\in\Pi, \Pi:=\{\mathbf{X}|\mathbf{X}_{ij}=\{0, 1\},\sum_{j=1}^N{\mathbf{X}_{ij}}=1,
\sum_{i=1}^M{\mathbf{X}_{ij}}\leq1,\forall i, j\}, M\leq N.
\end{equation}
Such a formulation covers a wide range of important problems, such as correspondence, assignment, matching, and traveling salesman problem (TSP).
Specific to graph matching, it defines a subgraph matching problem where each node in the smaller graph has to match exactly one node in the bigger
one, and each node in the bigger graph can match at most one node in the smaller one. It obviously includes (\ref{eqn_gm}) as a special case when
$M=N$.

To approximate the integer program (\ref{oriobj}) by a relaxation technique, $\Pi$ is firstly relaxed to its convex hull, i.e.,
the set of ($M\times N$) \emph{doubly sub-stochastic matrices} $\Omega$ \cite{Maciel03},
\begin{equation}
\Omega:=\{\mathbf{X}|\mathbf{X}_{ij}\geq 0,\sum_{j=1}^N{\mathbf{X}_{ij}}=1, \sum_{i=1}^M{\mathbf{X}_{ij}}\leq1,\forall i, j\}.
\end{equation}
Then, we propose the Graduated NonConvexity and Graduated Concavity Procedure (GNCGCP) to approximately solve it as follows,
\begin{equation}\label{GNCGCP}
F_{\zeta}(\mathbf{X})=\begin{cases}
(1-\zeta)F(\mathbf{X})+\zeta\mathrm{tr}\mathbf{X}^{\top}\mathbf{X} & \text{ if } 1\geq \zeta\geq 0,\\
(1+\zeta)F(\mathbf{X})+\zeta\mathrm{tr}\mathbf{X}^{\top}\mathbf{X} & \text{ if } 0> \zeta\geq -1,
\end{cases}, \mathbf{X}\in\Omega.
\end{equation}
In GNCGCP, $\zeta$ decreases gradually from 1 to -1, implying that the objective function $F_{\zeta}(\mathbf{X})$ becomes
gradually from $\mathrm{tr}\mathbf{X}^{\top}\mathbf{X}$ to $F(\mathbf{X})$ (graduated nonconvexity) and finally to
$-\mathrm{tr}\mathbf{X}^{\top}\mathbf{X}$ (graduated concavity). For each currently fixed $\zeta$, $F_{\zeta}(\mathbf{X})$ is
minimized by the Frank-Wolfe algorithm \cite{Frank56}, using the minimum of the previous $F_{\zeta}(\mathbf{X})$ as the starting
point. Here and hereafter, $F(\mathbf{X})$ is assumed to be neither convex nor concave, or otherwise, (\ref{GNCGCP}) is further
simplified accordingly. That is, the equation on $1\geq \zeta\geq 0$ ($0> \zeta\geq -1$) is removed in case $F(\mathbf{X})$
itself is convex (concave).


\begin{figure}
\begin{algorithmic}[1]

\State $\zeta\leftarrow 1, \mathbf{X}\leftarrow \mathbf{X}^0$

\While{$\zeta > -1 \wedge \mathbf{X}\notin \Pi$}

\While{$\mathbf{X}$ not converged} \Comment{Frank-Wolfe algorithm}

\State $\mathbf{Y}=\arg\min_\mathbf{Y} \mathrm{tr}\nabla F_{\zeta}(\mathbf{X})^{\top} \mathbf{Y},   \mathrm{ s.t.  }  \mathbf{Y}\in \Omega$

\State $\alpha=\arg\min_{\alpha} F_{\zeta}(\mathbf{X}+\alpha(\mathbf{Y}-\mathbf{X})), s.t.  \hspace{0.2cm}0\leq\alpha\leq 1$
\Comment{line search}

\State $\mathbf{X}\leftarrow \mathbf{X}+\alpha (\mathbf{Y}-\mathbf{X})$

\EndWhile

\State $\zeta \leftarrow \zeta - d\zeta$

\EndWhile\label{euclidendwhile}

\State \textbf{return} $\mathbf{X}$

\end{algorithmic}
\caption{Algorithmic framework of GNCGCP.}\label{GNCGCPAlg}
\end{figure}

The algorithmic framework of GNCGCP is given by the algorithm in Figure \ref{GNCGCPAlg}. In the algorithm, the gradient $\nabla
F_{\zeta}(\mathbf{X})$ takes the form
\begin{equation}\label{nablaF}
\nabla F_{\zeta}(\mathbf{X})=\begin{cases}
(1-\zeta)\nabla F(\mathbf{X})+2\zeta \mathbf{X} & \text{ if } 1\geq \zeta\geq 0,\\
(1+\zeta)\nabla F(\mathbf{X})+2\zeta \mathbf{X} & \text{ if } 0> \zeta\geq -1.
\end{cases}
\end{equation}
The linear program $\mathbf{Y}=\arg\min_\mathbf{Y} \mathrm{tr}\nabla F_{\zeta}(\mathbf{X})^{\top} \mathbf{Y},   \mathrm{ s.t.  }
\mathbf{Y}\in \Omega$ can be solved by the non-square Hungarian algorithm \cite{Bourgeois71}, and line search
$\alpha=\arg\min_{\alpha} F_{\zeta}(\mathbf{X}+\alpha(\mathbf{Y}-\mathbf{X}))$ can be efficiently solved by the backtracking
algorithm \cite{Boyd04}. The convergence of $\mathbf{X}$ is confirmed by checking whether
\begin{equation}\label{convergecond}
\mathrm{tr}\nabla F_{\zeta}(\mathbf{X})^{\top}(\mathbf{Y}-\mathbf{X})<\varepsilon|F_{\zeta}(\mathbf{X})+\mathrm{tr}\nabla
F_{\zeta}(\mathbf{X})^{\top}(\mathbf{Y}-\mathbf{X})|.
\end{equation}
Once $\mathbf{X}$ becomes discrete, the algorithm is terminated, even if $\zeta$ has not reached -1.

Without considering sparsity (of the adjacency matrix), storage complexity of the algorithm is $O(N^2)$, and the computational
complexity is roughly $O(N^3)$ resulting from matrix multiplication. Both complexities are comparable with those of the
(extended) Path following algorithm \cite{Zaslavskiy09,zyliu12}.
\subsection{Discussions and Interpretations}
\subsubsection{GNCGCP realizes a type of CCRP}
\begin{theorem}\label{theorem1}
GNCGCP (\ref{GNCGCP}) realizes a convex and concave relaxations procedure (\ref{ccrp}), with the convex and concave relaxations
respectively given by
\begin{eqnarray}\label{covrelax1}
F_v(\mathbf{X}) &=& F(\mathbf{X}) -\lambda_{min} \mathrm{tr}\left(\mathbf{X}^{\top}\mathbf{X}-\mathbf{JX}\right), \nonumber \\
F_c(\mathbf{X}) &=& F(\mathbf{X})-\lambda_{max} \mathrm{tr}\left(\mathbf{X}^{\top}\mathbf{X}-\mathbf{JX}\right),
\end{eqnarray}
where $\mathbf{J}:=\mathbf{1}_{N\times M}$ denotes the unit matrix consisting of all 1s, $\lambda_{min}$ and $\lambda_{max}$ denote the minimal and
maximal eigenvlaues of the Hessian matrix of $F(\mathbf{X})$, respectively.
\end{theorem}

To prove Theorem \ref{theorem1}, we derive $F_{\zeta}(\mathbf{X})$ by adding a constant $-\zeta M$ as follows,
\begin{eqnarray}
\arg\min_\mathbf{X} F_{\zeta}(\mathbf{X}) = \arg\min_\mathbf{X} \left[F_{\zeta}(\mathbf{X}) -\zeta M\right] =\arg\min_\mathbf{X}
\left[F_{\zeta}(\mathbf{X}) -\zeta\mathrm{tr}\mathbf{JX} \right], \mathbf{X}\in\Omega.
\end{eqnarray}
Therefore, $F_{\zeta}(\mathbf{X})$ can be equivalently rewritten by $\hat{F}_{\zeta}(\mathbf{X})$ as
\begin{eqnarray}\label{GNCGCP2}
\hat{F}_{\zeta}(\mathbf{X}) = F_{\zeta}(\mathbf{X}) -\zeta\mathrm{tr}\mathbf{JX} =\begin{cases}
(1-\zeta)F(\mathbf{X})+\zeta\mathrm{tr}\left(\mathbf{X}^{\top}\mathbf{X}-\mathbf{JX}\right) & \text{ if } 1\geq \zeta\geq 0,\\
(1+\zeta)F(\mathbf{X})+\zeta\mathrm{tr}\left(\mathbf{X}^{\top}\mathbf{X}-\mathbf{JX}\right) & \text{ if } 0> \zeta\geq -1.
\end{cases}
\end{eqnarray}
To prove that (\ref{GNCGCP2}) realizes exactly a CCRP, two Propositions were firstly given as follows.
\begin{proposition}\label{propo1}
There always exists a $\zeta_u=\frac{\lambda_{min}}{\lambda_{min}-1}\in(0, 1)$ making $\hat{F}_{\zeta}(\mathbf{X})$ convex as $1\geq
\zeta\geq\zeta_u$, where $\lambda_{min}$ denotes the smallest eigenvalue of the Hessian matrix $\mathbf{H}_{\mathbf{X}}$ of $F(\mathbf{X})$.
\end{proposition}
\emph{Proof: }The Hessian matrix $\hat{\mathbf{H}}_{\mathbf{X}}$ of $\hat{F}_{\zeta}(\mathbf{X})$ takes the form $(1-\zeta)\mathbf{H}_{\mathbf{X}}
+\zeta \mathbf{I}$. To make $\hat{\mathbf{H}}_{\mathbf{X}}$  positive definite, $\zeta$ should satisfy $\zeta\geq
\frac{\lambda_{min}}{\lambda_{min}-1}$. As $F(\mathbf{X})$ is neither convex nor concave, $\lambda_{min}$ is a negative number, which makes $0 <
\frac{\lambda_{min}}{\lambda_{min}-1} < 1$. Thus, choosing $\zeta_u=\frac{\lambda_{min}}{\lambda_{min}-1}$, any $\zeta$ satisfying $1\geq
\zeta\geq\zeta_u$ will make $\hat{\mathbf{H}}_{\mathbf{X}}$  positive definite and consequently $\hat{F}_{\zeta}(\mathbf{X})$ convex.$\Box$
\begin{proposition}\label{propo2}
There always exists a $\zeta_l=\frac{-\lambda_{max}}{\lambda_{max}+1}\in(-1, 0)$ making $\hat{F}_{\zeta}(\mathbf{X})$ concave as
$\zeta_l\geq\zeta\geq -1$, where $\lambda_{max}$ denotes the biggest eigenvalue of the Hessian matrix of $F(\mathbf{X})$.
\end{proposition}
\emph{Proof: }The proof can be accomplished in a similar way as that of Proposition \ref{propo1}.$\Box$

Then, based on the above two Propositions, we get the following two Lemmas.
\begin{lemma}\label{lemma1}
The value range $1\geq\zeta\geq0$ in (\ref{GNCGCP2}) can be equivalently shrunk to $\zeta_u\geq\zeta\geq0$ with $\zeta_u$ given
by Proposition \ref{propo1}.
\end{lemma}
\emph{Proof:} $\hat{F}_{\zeta_u}(\mathbf{X})$ is a convex function, whose global minimum is obtainable without depending on the previous results
gotten on $1\geq\zeta>\zeta_u$. Thus, the value range of $1\geq\zeta\geq0$ can be equivalently shrunk to $\zeta_u\geq\zeta\geq0$  for
(\ref{GNCGCP2}).$\Box$
\begin{lemma}\label{lemma2}
The value range $0>\zeta\geq-1$ in (\ref{GNCGCP2}) can be equivalently shrunk to $0>\zeta\geq\zeta_l$ with $\zeta_l$ given by
Proposition \ref{propo2}.
\end{lemma}
\emph{Proof: }$\hat{F}_{\zeta_l}(\mathbf{X})$ is a concave function, implying that minimization of $\hat{F}_{\zeta_l}(\mathbf{X})$
will result in a discrete solution $\mathbf{\hat{X}}\in\Pi$. As $\zeta$ decreases further from $\zeta_l$ to $-1$, $\mathbf{\hat{X}}$
will keep unchanged because it remains to be a local minimum of $\hat{F}_{\zeta}(\mathbf{X})$. Thus, the value range of $0>\zeta\geq-1$ can be equivalently shrunk to $0>\zeta\geq\zeta_l$ for  (\ref{GNCGCP2}).$\Box$\\
\emph{Comments:} Actually, once $\zeta$ reaches $\zeta_l$, the GNCGCP will terminate according to the algorithm in Figure
\ref{GNCGCPAlg}. Therefore, Lemma \ref{lemma2} holds naturally in the context of GNCGCP.

Finally, we prove Theorem \ref{theorem1}.

\emph{Proof of Theorem \ref{theorem1}: }Based on Lemmas \ref{lemma1} and \ref{lemma2}, $\hat{F}_{\zeta}(\mathbf{X})$ in (\ref{GNCGCP2}) (or
$F_{\zeta}(\mathbf{X})$ in (\ref{GNCGCP})) is equivalently rewritten as
\begin{eqnarray*}
\hat{F}_{\zeta}(\mathbf{X})= \begin{cases}
(1-\zeta)F(\mathbf{X})+\zeta\mathrm{tr}\left(\mathbf{X}^{\top}\mathbf{X}-\mathbf{JX}\right) & \text{ if } \zeta_u\geq \zeta\geq 0,\\
(1+\zeta)F(\mathbf{X})+\zeta\mathrm{tr}\left(\mathbf{X}^{\top}\mathbf{X}-\mathbf{JX}\right) & \text{ if } 0>\zeta\geq \zeta_l.
\end{cases}\\
\end{eqnarray*}
Then, for each fixed $\zeta$, $\hat{F}_{\zeta}(\mathbf{X})$ is normalized by a constant $1-\zeta$ or $1+\zeta$, making
\begin{eqnarray*}
\hat{F}_{\zeta}(\mathbf{X})=\begin{cases}
F(\mathbf{X})+\frac{\zeta}{1-\zeta}\mathrm{tr}\left(\mathbf{X}^{\top}\mathbf{X}-\mathbf{JX}\right) & \text{ if } \zeta_u\geq \zeta\geq 0,\\
F(\mathbf{X})+\frac{\zeta}{1+\zeta}\mathrm{tr}\left(\mathbf{X}^{\top}\mathbf{X}-\mathbf{JX}\right) & \text{ if } 0>\zeta\geq
\zeta_l,
\end{cases}
\end{eqnarray*}
or equivalently,
\begin{equation}\label{GNCGCP3}
\hat{F}_{\gamma}(\mathbf{X})= F(\mathbf{X})+\gamma\mathrm{tr}\left(\mathbf{X}^{\top}\mathbf{X}-\mathbf{JX}\right), -\lambda_{min}
\geq \gamma\geq -\lambda_{max},
\end{equation}
where $\lambda_{min}$ and $\lambda_{max}$ are defined in propositions \ref{propo1} and \ref{propo2}, respectively.

On the other hand, based on the convex and concave relaxations given by (\ref{covrelax1}), a CCRP is constructed as follows,
\begin{equation}
F_{\eta}(\mathbf{X}) = (1-\eta)F_v(\mathbf{X})+\eta F_c(\mathbf{X})=F(\mathbf{X})-\left[(1-\eta)\lambda_{min} +
\eta\lambda_{max}\right]\mathrm{tr}\left(\mathbf{X}^{\top}\mathbf{X}-\mathbf{JX}\right).
\end{equation}
By defining $\gamma=(1-\eta)\lambda_{min} + \eta\lambda_{max}, 0\leq\eta\leq 1$, $F_{\eta}(\mathbf{X})$ above can be equivalently
written as
\begin{equation}
F_{\gamma}(\mathbf{X})= F(\mathbf{X})+\gamma\mathrm{tr}\left(\mathbf{X}^{\top}\mathbf{X}-\mathbf{JX}\right), -\lambda_{min} \geq
\gamma\geq -\lambda_{max}.
\end{equation}
Exact (\ref{GNCGCP3})! Therefore, the GNCGCP realizes a CCRP with the convex and concave relaxations given by (\ref{covrelax1}).$\Box$

It is worth discussing the case of $F(\mathbf{X})$ being non-quadratic (such as a quartic function like (\ref{subgm-obj})) where
$\lambda_{min}$ and $\lambda_{max}$ are in general dependent on $\mathbf{X}$. If $\mathbf{X}$ is unconstrained/unbounded,
$\lambda_{min}$ and $\lambda_{max}$ might become $-\infty$ and $+\infty$ respectively, implying that any $-1<\zeta<1$ will result
in $F_{\zeta}(X)$ neither convex nor concave (see Propositions \ref{propo1} and \ref{propo2}). Fortunately, because $\mathbf{X}$
here is constrained as a doubly sub-stochastic matrix, i.e., each element is bounded by $0\leq x\leq 1$, both $\lambda_{min}$ and
$\lambda_{max}$ must be some finite numbers, meaning that we can always get a convex relaxation by some $\zeta < 1$ and a concave
relaxation by some $\zeta>-1$. The point here is that GNCGCP does not need to figure out the number ($\lambda$ or corresponding
$\zeta$) explicitly, which is realized in an implicit way.

\begin{figure}[h]
\centering
   \scalebox{.6}[.5]{\includegraphics[scale=.8]{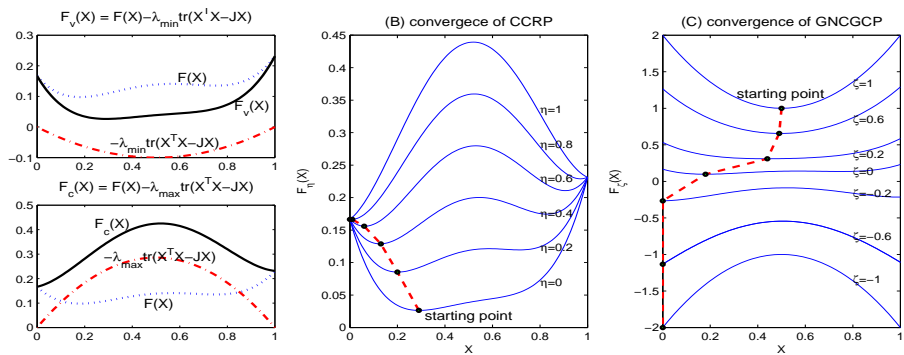}}
  \caption{Illustration of the construction of convex and concave relaxations (the one on the left-hand side), and the convergence processes of CCRP and GNCGCP.}\label{Fig:convergence}
\end{figure}

A simple illustration of the convex and concave relaxations in (\ref{covrelax1}) is shown in Figure \ref{Fig:convergence} (the sub-figure on the
left-hand side).

\subsubsection{GNCGCP versus CCRP}\label{sec_discussion} Basically, without involving convex or concave relaxation explicitly, GNCGCP provides a
very simple way to construct a CCRP algorithm; by contrast, the problem specific relaxation is typically difficult to construct. A typical example is
the complicated concave relaxations used by (E)PATH \cite{Zaslavskiy09,zyliu12}, which are applicable only on equal-sized graph matching. Similarly,
it is also usually difficult to calculate $\lambda_{max}$ or $\lambda_{min}$ in (\ref{covrelax1}), especially on non-quadratic functions.

Another interesting difference between GNCGCP and CCRP lies in the construction of relaxation functions. A convex or concave relaxation will
certainly reshape the original relaxed function when $\mathbf{X}\in\Omega$ or $\in\mathcal{D}$. By introducing the simple quadratic function
$\zeta\mathrm{tr}\mathbf{X}^{\top}\mathbf{X}$, GNCGCP reshapes the relaxed function in a symmetric way, and meanwhile, as $\zeta\rightarrow 0$,
GNCGCP approaches the original relaxed function. By contrast, other types of relaxations in general have no chance to directly optimize it. This is
probably the main reason GNCGCP exhibited a better or at least a no worse performance than some other types of CCRP, especially on the QAP discussed
in Section \ref{s4}.

A simple comparison between the convergence of CCRP and GNCGCP is given by Figure \ref{Fig:convergence}, where it is observed
that CCRP starts with a convex relaxation and ends with a concave relaxation but GNCGCP starts with
$\mathrm{tr}\mathbf{X}^{\top}\mathbf{X}$ and ends with $-\mathrm{tr}\mathbf{X}^{\top}\mathbf{X}$, with the convex and concave
relaxations realized implicitly during the process (see Lemmas \ref{lemma1} and \ref{lemma2}).

Based on the GNCGCP algorithm in Figure \ref{GNCGCPAlg}, to utilize GNCGCP the only thing we need to do is to find the gradient
of the objective function. Thus, any optimization problems on $\Pi$ with a differentiable objective function can be directly
approximated by GNCGCP. Below we use (sub)graph matching and quadratic assignment problem to demonstrate this simple process.

\section{Case Study 1: (Sub)Graph Matching}\label{s3}
\subsection{Problem Formulation}
(Sub)Graph matching as a fundamental problem in theoretical computer science finds wide applications in computer vision and machine
learning \cite{Umeyama88,Gold96,Conte04,Leordeanu_2009_6508,Zaslavskiy09,zyliu12,Zhou12}. Given two graphs $G_D$ and $G_M$ to be matched, the
(sub)graph matching problem is formulated as follows,
\begin{equation}\label{subgm-obj}
\min_\mathbf{X} F(\mathbf{X})=\mathrm{tr}(\mathbf{A}_{M}-\mathbf{XA}_D\mathbf{X}^{\top})^{\top}(\mathbf{A}_{M}-\mathbf{XA}_D\mathbf{X}^{\top}),
\mathrm{s.t.} \mathbf{X}\in\Pi,
\end{equation}
where $\mathbf{A}_M\in\mathbb{R}^{N_M\times N_M}$ and $\mathbf{A}_D\in\mathbb{R}^{N_D\times N_D}$ denote the adjacency matrices of $G_M$ and $G_D$,
respectively, and $N_M \leq N_D$. To use GNCGCP to approximate it, we need just to relax $\Pi$ to $\Omega$ and find $\nabla F(\mathbf{X})$ as,
\begin{equation}
\nabla F(\mathbf{X})
=2\mathbf{X}(\mathbf{A}_D^{\top}\mathbf{X}^{\top}\mathbf{XA}_D+\mathbf{A}_D\mathbf{X}^{\top}\mathbf{XA}_D^{\top})-2(\mathbf{A}_M\mathbf{XA}_D^{\top}+\mathbf{A}_M^{\top}\mathbf{XA}_D).
\end{equation}
Below we denote by GNCGCP\_SGM the above (sub)graph matching algorithm, which is applicable on both equal-sized and subgraph
matching problems, and is applicable on any types of graph provided that it can be represented by an adjacency matrix.

In case the two graphs take exactly the same size $N$ which implies that $\Pi$ degenerates to $\mathcal{P}$, the objective function in
(\ref{subgm-obj}) can be derived as \cite{Zaslavskiy09},
\begin{eqnarray}\label{gm_obj}
F(\mathbf{X})=\mathrm{tr}(\mathbf{A}_{M}-\mathbf{XA}_D\mathbf{X}^{\top})^{\top}(\mathbf{A}_{M}-\mathbf{XA}_D\mathbf{X}^{\top})=\mathrm{tr}(\mathbf{A}_{M}\mathbf{X}-\mathbf{XA}_D)^{\top}(\mathbf{A}_{M}\mathbf{X}-\mathbf{XA}_D)
, \mathbf{X}\in\mathcal{P}.
\end{eqnarray}
Then, by relaxing $\mathcal{P}$ to $\mathcal{D}$, GNGGCP is implementable by finding
\begin{equation}
\nabla F(\mathbf{X}) = \mathbf{A}_M^{\top}\mathbf{A}_M\mathbf{X} - \mathbf{A}_M^{\top}\mathbf{X} \mathbf{A}_D -
\mathbf{A}_M\mathbf{X}\mathbf{A}_D^{\top} + \mathbf{XA}_D\mathbf{A}_D^{\top}.
\end{equation}
Because $F(\mathbf{X})$ with $\mathbf{X}\in \mathcal{D}$ in (\ref{gm_obj}) itself becomes a convex function, GNCGCP is further
simplified by removing the equation on $1\geq \zeta\geq 0$, that is, $\zeta$ needs just to decrease from $0$ to $-1$ but not $1$
to $-1$. The algorithm is denoted by GNCGCP\_GM, which is closely related to the Path following \cite{Zaslavskiy09} (on
undirected graph) and extended Path following \cite{zyliu12} algorithms, with the same convex relaxation but a different concave
relaxation.

\subsection{Experimental Results}
\subsubsection{overview} Both synthetic and real data were employed to evaluate the GNCGCP algorithms.

On equal-sized graph matching, six algorithms including 1:) Umeyama's spectral decomposition (U for short) \cite{Umeyama88}, 2:) graduated assignment
(GA) \cite{Gold96}, 3:) path following algorithm (PATH, for undirected graph only) \cite{Zaslavskiy09}, 4:) extended path following (EPATH, for
directed graph only)  \cite{zyliu12}, 5:) GNCGCP\_SGM, and 6:) GNCGCP\_GM were experimentally compared. Considering space limit we are not to compare
their complexities in detail. Actually, in all the following experiments, the time-cost of GNCGCP is comparable with that of (E)PATH.

On subgraph matching, four algorithms including GNCGCP\_SGM, GA, spectral relaxation matching (SM for short) \cite{leordeanu2005spectral}, and
probabilistic spectral matching (PSM) \cite{egozi2013probabilistic} were experimentally compared.

All of the algorithms were implemented by Matlab \footnote{The source codes of all the (sub)graph matching and QAP algorithms used in the experiments
are available at http://www.escience.cn/people/zyliu/GNCGCP.html}, and for GNCGCP\_SGM, GNCGCP\_GM and (E)PATH, the same parameter settings were used
as follows: the learning step $d\zeta=d\eta=0.001$ and the stopping parameter $\varepsilon=0.001$ in (\ref{convergecond}).

\subsubsection{on synthetic data}
Synthetic graphs were generated according to three options:
\begin{itemize}
  \item directed (abbreviated by D) or undirected (U);
  \item \textbf{degree} distribution: a binomial (B)($P(k)=C_N^kp^k(1-p)^{1-k}$ (with $p=0.5$) or a power (P) ($P(k)\propto k^{-\alpha}$) law
(scale-free graph with a fixed $\alpha=1.5$ in all of the experiments);
  \item \textbf{weight} distribution: a standard log-normal (L) ($p(w)=\frac{1}{w\sqrt{2\pi}}e^{-\frac{\ln^2 w }{2}},w>0$) or absolute normal
(N)($p(w)=\frac{2}{\sqrt{2\pi}}e^{-\frac{w^2}{2}},w\geq 0$).
\end{itemize}

Therefore, there are totally eight types of graphs, each of which is abbreviated by a sequential three-character notation. For instance, DBL denotes
the directed graphs with a binomial degree distribution and a log-normal weight distribution.

\begin{figure}[h]
   \hspace{-2.5cm}\centering \scalebox{.55}[.5]{\includegraphics[scale=1]{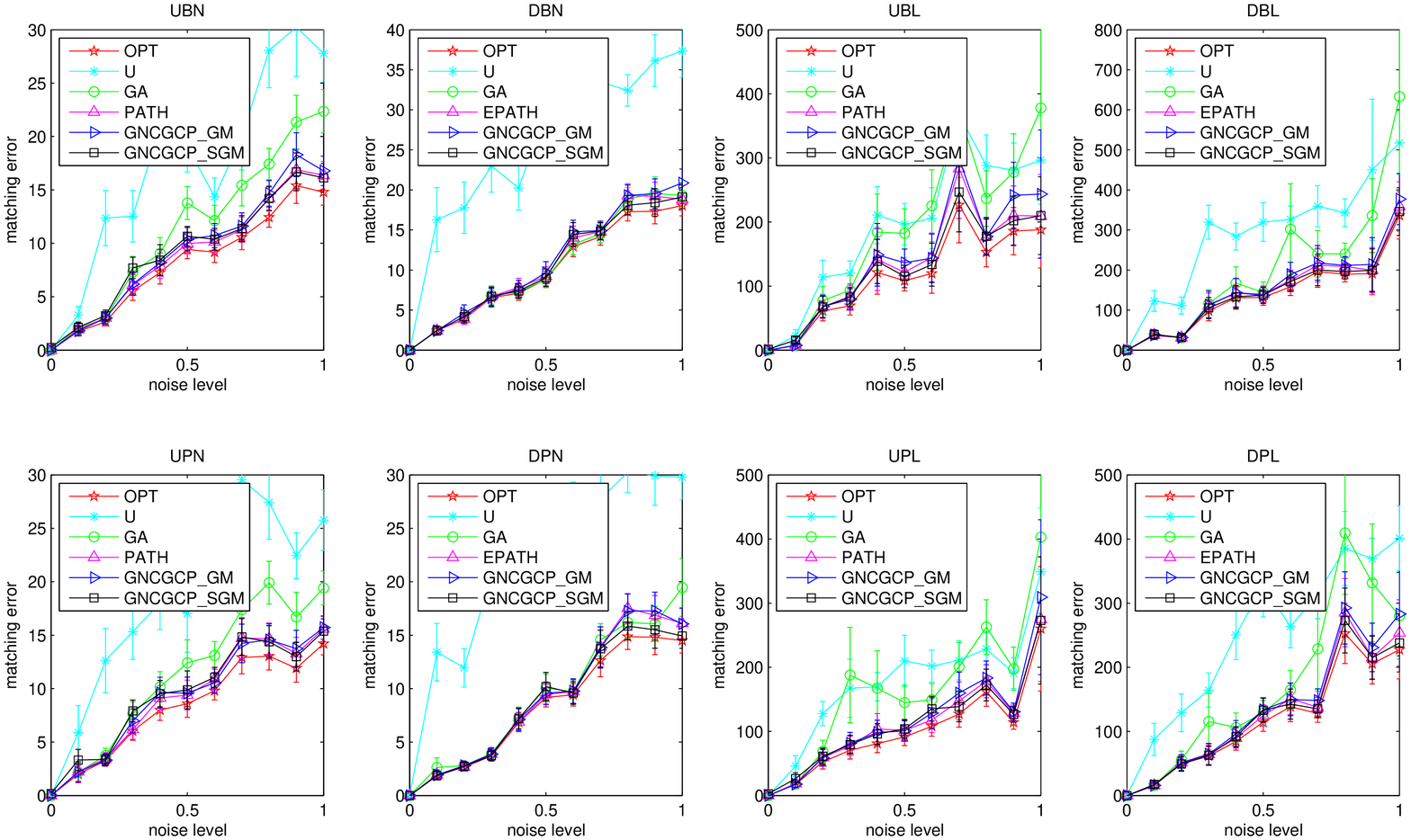}}
  \caption{Matching errors on the eight types of graphs with respect to noise levels,
summarized from 50 random runs on each noise level.  }\label{Fig:similar}
\end{figure}

Two experiments were conducted on equal-sized graphs, with the first one to evaluate the noise resistance ability of the
algorithms, and the second one to evaluate their scalabilities with respect to graph size. In the first experiment, the graph
size $N$ was fixed at 8, and for each graph pair, $G_M$ was generated by adding $\beta |E_D|$ edges into $G_D$, where $|E_D|$
denotes the number of edges of $G_D$, and $\beta$ is the parameter that controls the noise level. In the experiment $\beta$ was
increased from 0 to 1 by a step size $0.1$, and on each noise level, 50 graph pairs were randomly generated. The experimental
results on the eight types of graphs are shown in Figure \ref{Fig:similar}, where OPT denotes the optimal result obtained by an
exhaustive search.

\begin{figure}[h]
\centering
  \hspace{-2.5cm} \scalebox{.55}[.5]{\includegraphics[scale=1]{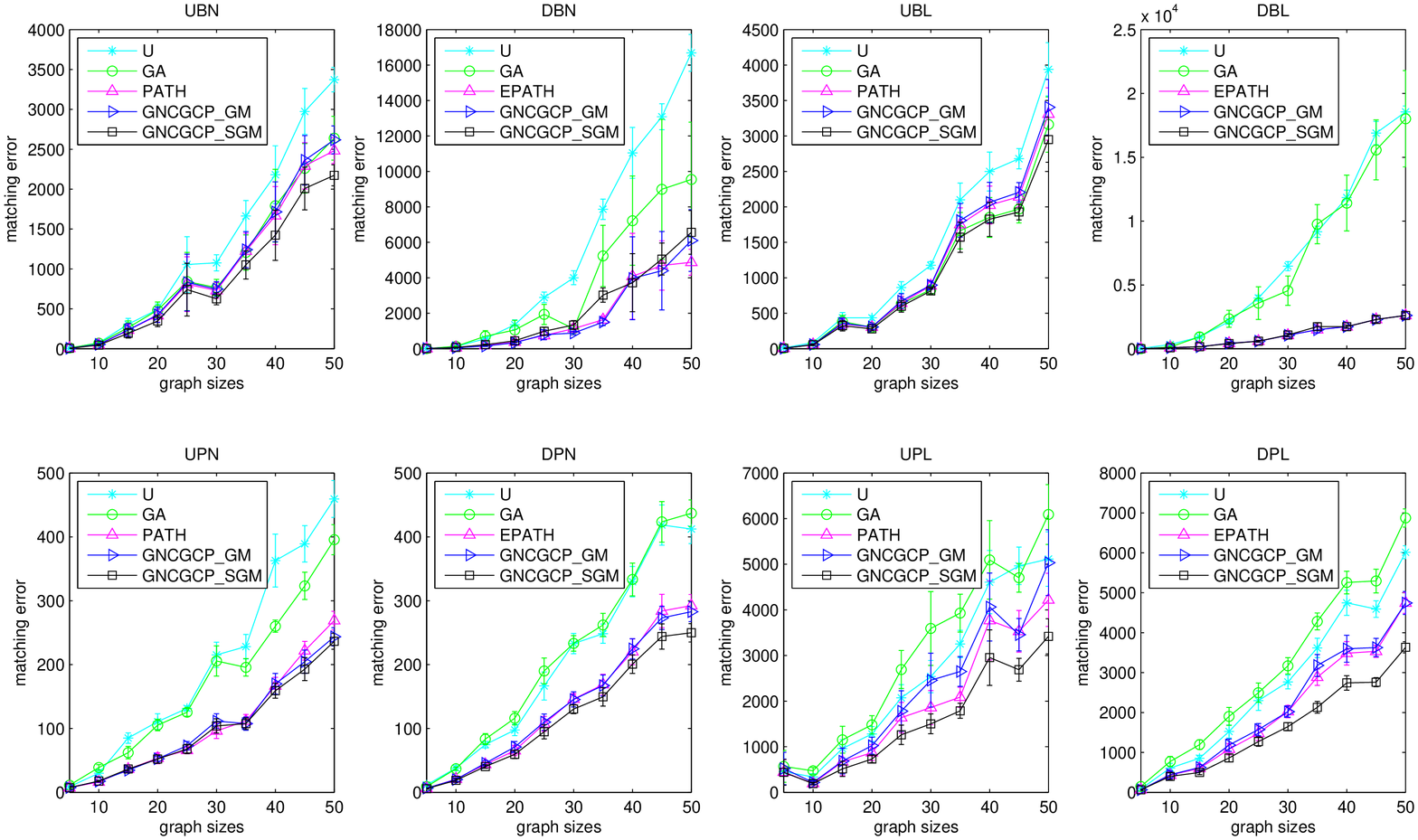}}
  \caption{Matching errors on the eight types of graphs with respect to graph sizes,
summarized from 50 random runs on each size. }\label{Fig:scale}
\end{figure}

In the second experiment, 10 groups of graph pairs were generated for each of the eight types, with the graph size increasing from 5 to 50 by a step
size 5. For each group 50 graph pairs were randomly generated in the same way as the first experiment with a fixed noise level 0.2. The experimental
results are shown in Figure \ref{Fig:scale}.

Two observations could be summarized from the above experimental results. First, (E)PATH, GNCGCP\_GM and GNCGCP\_SGM outperformed significantly U and
GA. This witnessed the superiority of CCRP and also GNCGCP. Second, GNCGCP\_SGM exhibited a slightly better or at least a no worse performance than
GNCGCP\_GM and (E)PATH (see for instance UPL and DPL in Figure \ref{Fig:scale}), and meanwhile GNCGCP\_GM exhibited a comparable performance with
(E)PATH, echoed by the discussions in Section \ref{sec_discussion}.

\begin{figure}[h]
\centering \scalebox{.6}[.5]{\includegraphics[scale=1]{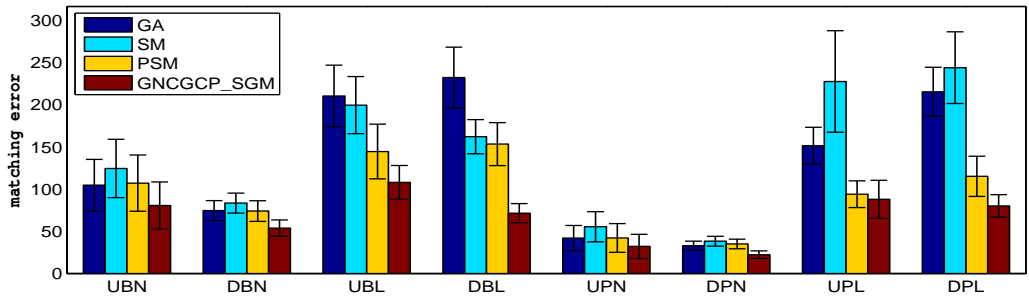}}
  \caption{Subgraph matching results on the eight types of synthetic graphs, summarized from 50 random runs on each type. }\label{Fig:subgraph}
\end{figure}

On subgraph matching, $G_D$ and $G_M$ were generated in the following way. The bigger graph $G_D$ with $N_D=20$ was firstly randomly generated, then
a smaller graph $G_{\hat{M}}$ with $N_M=10$ was randomly extracted out from $G_D$, and finally $G_M$ was generated by adding some noises to
$G_{\hat{M}}$ in the same manner as the first experiment by setting $\beta = 0.5$. The experimental results are shown in Figure \ref{Fig:subgraph},
where GNCGCP\_SGM achieved the best performance on all of the eight types of graphs.

\subsubsection{on real data}
\begin{figure}[h]
\centering \hspace{-3cm}\scalebox{1}[1]{\includegraphics[width=10cm, bb = 0 388 461 471]{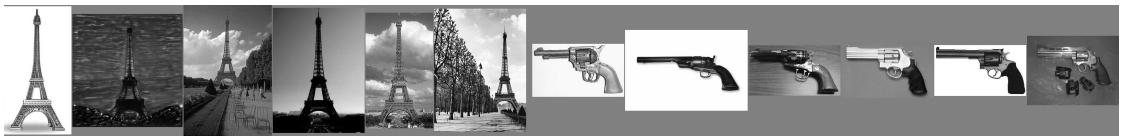}}
  \caption{The \emph{eiffel} and \emph{revolver} images used in the experiments.}\label{Fig:realdata}
\end{figure}

The real data experiments were conducted on the 6 \emph{eiffel} and 6 \emph{revolver} samples shown in Figure \ref{Fig:realdata}, which were fetched
from the Caltech-256 Database \cite{cal256}. The first one and first five leftmost samples of \emph{eiffel} and \emph{revolver} in Figure
\ref{Fig:realdata} were chosen as the model samples to match the rest five and one samples, respectively. Total 35 and 20 feature points (typically
corner points) were marked manually for the \emph{eiffel} and \emph{revolver} models respectively, and all of the points are linked each other to
construct their undirected graph representation. The edge attribute comprises two parts, i.e., the normalized $distance$ and $direction$, and the
unary term or appearance cue was not incorporated into the model.

\begin{figure}[h]
\centering \scalebox{.5}[.35]{\includegraphics[scale=1]{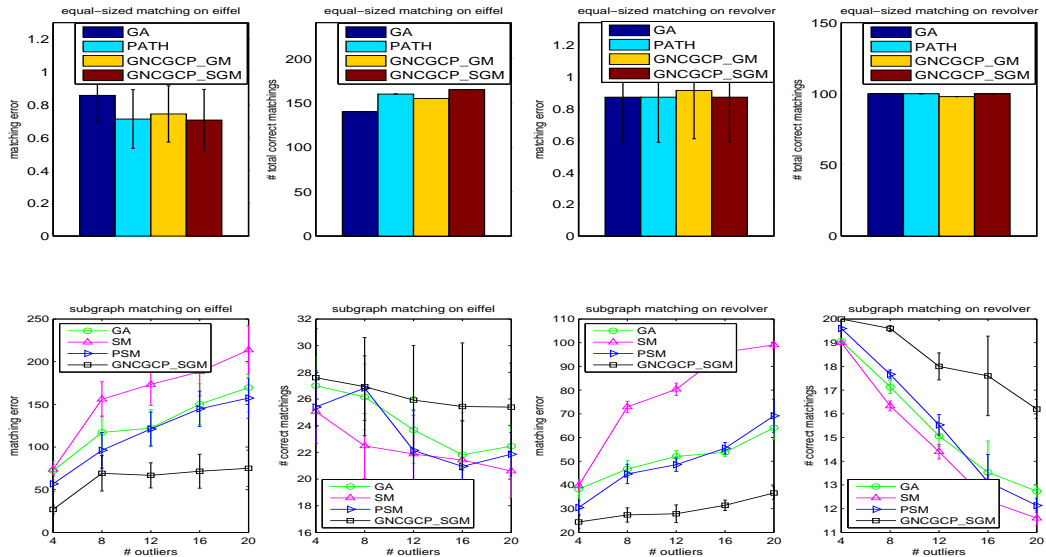}}
  \caption{The experimental results on \emph{eiffel} and \emph{revolver} images, where both the matching error and number of corrected matchings are presented. The upper row is on equal-sized matching and lower row is on subgraph matching.}\label{Fig:realresults}
\end{figure}

\begin{figure}
  \centering
  \subfigure[On \emph{eiffel} images]{
    \label{fig:subfig:a} 
    \includegraphics[width=7.5cm, bb = 57 323 581 542]{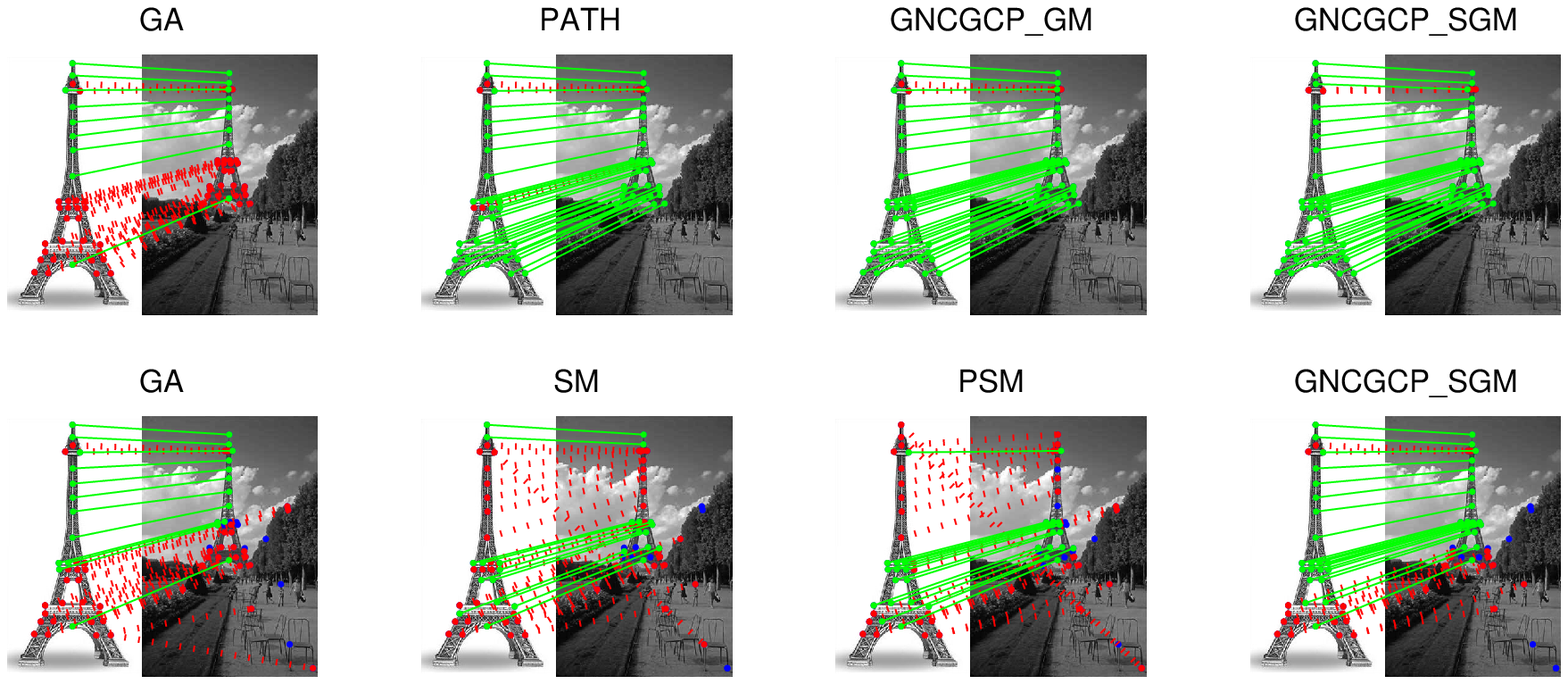}}
    \hspace{.5cm}
  \subfigure[On \emph{revolver} images]{
    \label{fig:subfig:a} 
    \includegraphics[width=7.5cm, bb = 1 330 595 539]{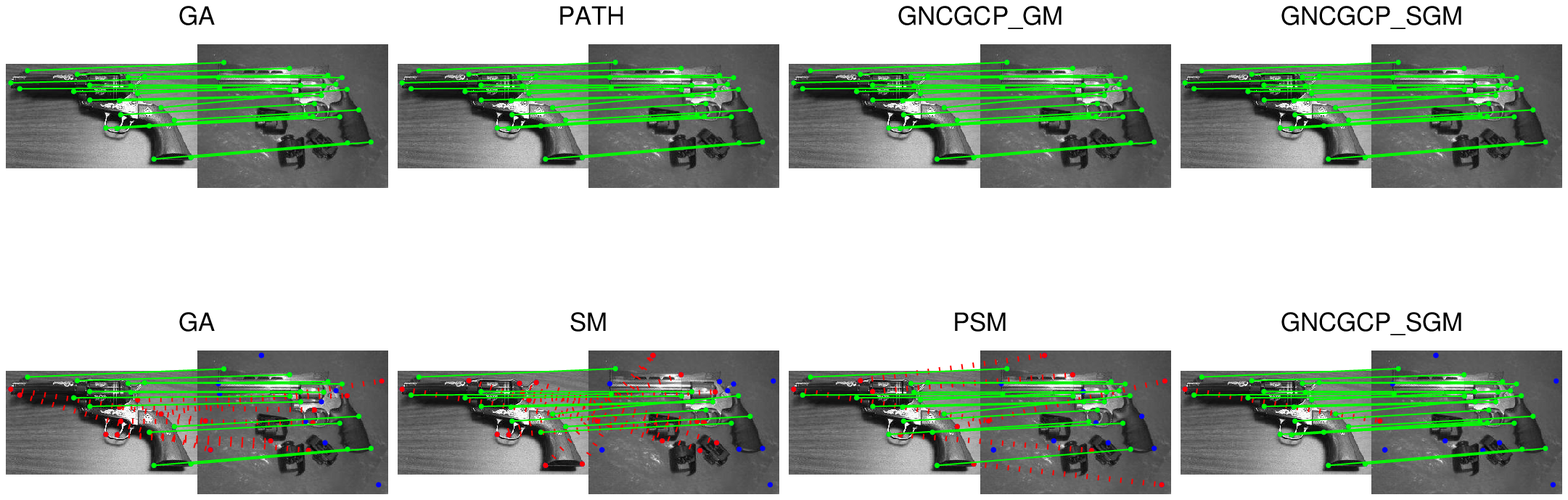}}
  \caption{Some typical equal-sized and subgraph matching results. The upper row is on equal-sized matching and lower row is on subgraph matching with 12 outliers, where the dot lines denote
  wrong matchings.}
  \label{fig:subfig} 
\end{figure}

The equal-sized graph matching results are shown in Figure \ref{Fig:realresults} (the upper row), including both the matching error and summed number
of correct matchings (whole numbers are 175 and 100 for \emph{eiffel} and \emph{revolver} respectively), and some typical matchings are shown in
Figure \ref{fig:subfig} (the upper row). It is observed that on \emph{eiffel} GNCGCP\_SGM got slightly better results, and on \emph{revolver} all the
four algorithms got quite good results, with the number of correct matchings being 100, 100, 98, and 100 respectively.

We then conducted subgraph matching on the data, by taking the one \emph{eiffel} and five \emph{revolver} model samples as
smaller models, and randomly adding some outlier points to the rest samples to get the larger one. The algorithms were evaluated
on five levels of the number of added outliers, i.e., 4, 8, 12, 16, and 20 respectively. The experimental results are shown in
Figure \ref{Fig:realresults} (the lower row), and some typical results by adding 12 outliers are shown in Figure \ref{fig:subfig}
(the lower row). We can observe that GNCGCP\_GCP got the best results on both criterions, and meanwhile, though the performance
became in general worse as the number of outliers became bigger, the decline of GNCGCP\_SGM was the slowest one.


\section{Case Study 2: Quadratic Assignment Problem}\label{s4}
The quadratic assignment problem (QAP) is a well-known combinatorial optimization problem in operations research and discrete
optimization \cite{rangarajan96-2,Misevicius03,Loiola07}, and is closely related to the equal-sized graph matching problem. Given
two equal-sized matrices $\mathbf{A}, \mathbf{B}$, and without considering the linear term, QAP formally takes the following
form,
\begin{eqnarray}\label{eqn_qap}
\min_\mathbf{X} F(\mathbf{X}) = \mathrm{tr}(\mathbf{AXB}^{\top}\mathbf{X}^{\top}),\hspace{0.5cm} s.t. \mathbf{X}\in \mathcal{P}.
\end{eqnarray}
By relaxing $\mathcal{P}$ to $\mathcal{D}$, the GNCGCP is applicable by finding
\begin{equation}
\nabla F(\mathbf{X}) = \mathbf{AXB}^{\top}+\mathbf{A}^{\top}\mathbf{XB}.
\end{equation}
The algorithm is denoted by GNCGCP\_QAP.

GNCGCP\_QAP was compared with PATH and EPATH on both the symmetric and asymmetric QAPLib benchmark datasets \cite{Burkard97} used respectively in
\cite{Zaslavskiy09} and \cite{zyliu12}. GNCGCP\_SGM and GNCGCP\_GM were also included in the experiments. The parameter settings of different
algorithms were the same as those in the previous (sub)graph matching experiments. The experimental results are listed in Tables \ref{tab:tab_sym}
and \ref{tab:tab_asym} respectively, where the results except for the three types of GNCGCP algorithms are directly fetched from \cite{Zaslavskiy09}
and \cite{zyliu12} respectively, OPT denotes the currently known best result, and for each algorithm, 'awar(\%)' (average wrong assignment ratio) =
$\frac{1}{n}\sum_{i=1}^n(cost_i-opt_i)/opt_i$ indicates its average deviation from OPT.

\begin{table} [htb]
\begin{center}
\caption{Comparative results on some symmetric QAPLib benchmark data sets.} \label{tab:tab_sym}
\begin{tabular}{r|cccccccc} \hline
 Data &OPT& U & GA & QPB & PATH&GNCGCP\_QAP &GNCGCP\_SGM& GNCGCP\_GM\\ \hline
chr12c&11156& 40370 & 19014 &20306 &18048 &\textbf{11566}&17020&21818\\
chr15a&9896& 60986 & 30370 &26132 &19088&12402&\textbf{10840}&19186 \\
chr15c&9504& 76318 & 23686 & 29862 & 16206&15080 &\textbf{14890} &19942\\
chr20b&2298& 10022 & 6290 &6674 &5560&\textbf{3164}&3452&5286\\
chr22b&6194& 13118 & 9658 &9942 &8500&\textbf{6918}&7858&8358\\
rou12 &235528& 295752 & 273438 &278834 &256320 &\textbf{238134}&238954 & 264324\\
rou15 &354210& 480352 & 457908 &381016 &391270&\textbf{374932}&377898&391768\\
rou20 &725522& 905246 & 840120 &804676 &778284&\textbf{729542}& 747322& 772100\\
tai10a&135028& 189852 & 168096 &165364 &152534&\textbf{138306}& 138306& 147092\\
tai15a&388214& 483596 & 451164 &455778 &419224&\textbf{392268}& 396596& 419328\\
tai17a&491812& 620964 & 589814 &550852 &530978&\textbf{514224}& 531732 &545802\\
tai20a&703482& 915144 & 871480 &799790 &753712&\textbf{736710} &746766 &778154\\
tai30a&1818146& 2213846 & 2077958 &1996442 &1903872&\textbf{1856666}& 1874642 &1917674\\
tai35a&2422002& 2925390 & 2803456 &2720986 &2555110&\textbf{2470186}&2482652 & 2544354\\
tai40a&3139370& 3727478 & 3668044 &3529402 &3281830&\textbf{3180740}& 3224410& 3342272\\ \hline
awar(\%)&0&146.5 &56.6 & 56.5&32.1&\textbf{10.9}&15.6&34.7\\
\hline
\end{tabular}\\
\end{center}
\end{table}

\begin{table} [htb]
\begin{center}
\caption{Comparative results on some asymmetric QAPLib benchmark data sets.} \label{tab:tab_asym}
\begin{tabular}{r|cccccccc} \hline
 Data &OPT&U & QCV & GA & EPATH&GNCGCP\_QAP &GNCGCP\_SGM& GNCGCP\_GM\\ \hline
lipa20a&3683&3925 & 3902&3909  &3885& \textbf{3789} &3823 & 3860\\
lipa20b&27076& 35213 & 34827 &\textbf{27076} &32081& \textbf{27076}& \textbf{27076}& 32207\\
lipa30a&13178& 13841 & 13787 &13668 & 13577& \textbf{13459} & 13485& 13628\\
lipa30b&151426& 196088& 189496 &\textbf{151426} &\textbf{151426} &\textbf{151426}& \textbf{151426}&\textbf{151426}\\
lipa40a&31538& 32663 & 32647 &32590 &32247&32024& \textbf{32012}&32356 \\
lipa40b&476581& 626004 & 572039 &\textbf{476581} &\textbf{476581}&\textbf{476581}& \textbf{476581}&\textbf{476581} \\
lipa50a&62093& 64138 & 63930 &63730 &63339&\textbf{62901}&63032& 63414\\
lipa50b&1210244& 1569908 & 1468492 &\textbf{1210244} &\textbf{1210244}&\textbf{1210244}&\textbf{1210244} &\textbf{1210244}\\
lipa60a&107218& 110196 & 110075 &109809 &109168&\textbf{108445}&108679&109079 \\
lipa60b&2520135& 3305286 & 3131985 &\textbf{2520135} &\textbf{2520135}&\textbf{2520135}& \textbf{2520135}& \textbf{2520135}\\
lipa70a&169755& 173906 & 173496 &173172 &172200&\textbf{171421}&171723 & 172519\\
lipa70b&4603200& 5974833 & 5576103 &\textbf{4603200} &\textbf{4603200}&\textbf{4603200}&\textbf{4603200}&\textbf{4603200}\\
lipa80a&253195& 258262 & 258140 &258218 &256601&255639&\textbf{255546}&256430\\
lipa80b&7763962&10079359 & 9703626 &\textbf{7763962} &\textbf{7763962}&\textbf{7763962}&\textbf{7763962}&\textbf{7763962}\\
lipa90a&360630& 367756 & 367250 &366743 &365233&\textbf{363480}&364319&364900\\
lipa90b&12490441& 16271254 & 13870571 &\textbf{12490441} &\textbf{12490441}&\textbf{12490441}&\textbf{12490441}&\textbf{12490441}\\
 \hline
 awar(\%)&0&16.81 &12.62 & 1.48&2.30&\textbf{0.72}&0.86&2.36\\ \hline
 \end{tabular}\\
\end{center}
\end{table}

It is observed that GNCGCP\_QAP exhibited the best performance. Specifically, it achieved the best results on 27 out of the 31 datasets, and in
average, GNCGCP\_QAP outperformed all of the competitors on both the symmetric and asymmetric datasets. It is also interesting to observe that
GNCGCP\_SGM outperformed (or achieved the same best results on some asymmetric datasets) both (E)PATH and GNCGCP\_GM on all of the 31 datasets.

Actually, by setting $\mathbf{A}_M:=-\mathbf{A}^{\top}$ and $\mathbf{A}_D:=\mathbf{B}^{\top}$, to utilize GNCGCP\_SGM to solve (\ref{eqn_qap}), one
needs to add the term
$$F_{1}(\mathbf{X})=\mathrm{tr}(\mathbf{XA}_{D}^{\top}\mathbf{X}^{\top}\mathbf{XA}_{D}\mathbf{X}^{\top})$$
into (\ref{subgm-obj}) to get (\ref{eqn_qap}). Similarly, the term
$$F_{2}(\mathbf{X})=\mathrm{tr}(\mathbf{X}^{\top}\mathbf{A}_{M}^{\top}\mathbf{A}_{M}\mathbf{X})+\mathrm{tr}(\mathbf{A}_{D}^{\top}\mathbf{X}^{\top}\mathbf{X}\mathbf{A}_{D})$$
has to be added into (\ref{gm_obj}) to get (\ref{eqn_qap}). Both $F_1(\mathbf{X})$ and $F_2(\mathbf{X})$ become constant when
$\mathbf{X}\in\mathcal{P}$, implying that all of the GNCGCP\_SGM, GNCGCP\_GM and (E)PATH implement a CCRP algorithm to solve QAP. But when
$\mathbf{X}\in \mathcal{D}$, $F_1(\mathbf{X})$ and $F_2(\mathbf{X})$ will certainly reshape or provide some biases on the original relaxed function;
as discussed in Section \ref{sec_discussion}, this is probably the main reason GNCGCP\_QAP achieved the best results. Meanwhile, it seems that
$F_1(\mathbf{X})$ which involves only $\mathbf{A}_D$ has less impact than $F_2(\mathbf{X})$ which involves both $\mathbf{A}_M$ and $\mathbf{A}_D$,
and therefore GNCGCP\_SGM achieved better results than both (E)PATH and GNCGCP\_GM.

\section{Conclusions}\label{s5}
The GNCGCP is proposed as a general optimization framework for the discrete optimization problems on the set of partial
permutation matrices, including a wide range of classic discrete optimization problems as its special cases, matching,
assignment, and traveling salesman problem (TSP), to name a few. GNCGCP has its root in the CCRP, but it does not need to figure
out the convex or concave relaxation explicitly, and is thus very easy to use in practical applications. Two case studies on
(sub)graph matching and QAP witness the simplicity as well as state-of-the-art performance of GNCGCP.


\end{document}